\documentclass{article}
\usepackage{ijcai22}

\usepackage{times}
\usepackage{soul}
\usepackage{url}
\usepackage[hidelinks]{hyperref}
\usepackage[utf8]{inputenc}
\usepackage[small]{caption}
\usepackage{graphicx}
\usepackage{amsmath}
\usepackage{amsthm}
\usepackage{booktabs}
\usepackage{algorithm}
\usepackage{algorithmic}
\usepackage{bbm}
\usepackage{multirow}
\usepackage{bbding}
\usepackage{pifont}
\usepackage{wasysym}
\usepackage{amssymb}
\usepackage{bbding}
\usepackage{caption}
 \usepackage{enumitem}

\title{
Lifelong Learning and Selective Forgetting via Contrastive Strategy
}

\author{
Lianlei Shan$^1$
\and
Wenzhang Zhou$^2$\and
Wei Li$^{2,3}$\And
Xingyu Ding$^4$
\affiliations
$^1$University of Chinese Academy of Sciences\\
$^2$Nanjing University of Posts and Telecommunications\\
$^3$Beijing University of Posts and Telecommunications\\
$^4$University of Chinese Academy of Sciences
\emails
shanlianlei18@mails.ucas.edu.cn,
0230196@njupt.edu.cn,
leesoon@bupt.edu.cn,
dingxingyu21@mails.ucas.ac.cn
}
\begin{document}

\maketitle

\begin{abstract}
Lifelong learning aims to train a model with good performance for new tasks while retaining the capacity of previous tasks. However, some practical scenarios require the system to forget undesirable knowledge due to privacy issues, which is called selective forgetting. The joint task of the two is dubbed Learning with Selective Forgetting (LSF). In this paper, we propose a new framework based on contrastive strategy for LSF. Specifically, for the preserved classes (tasks), we make features extracted from different samples within a same class compacted. And for the deleted classes, we make the features from different samples of a same class dispersed and irregular, i.e., the network does not have any regular response to samples from a specific deleted class as if the network has no training at all.
% 这种通过保持和打乱特征分布的方式可以使得不同类别的遗忘和记忆互不干扰。
Through maintaining or disturbing the feature distribution, the forgetting and memory of different classes can be or independent of each other.
% This approach guarantees precise memory for preserved classes, while achieving rapid and complete forgetting for deleted classes.
% Moreover, the previous settings of LSF are different from lifelong learning, and we unify them so that the off-the-shelf lifelong learning approaches can be easily modified to achieve selective forgetting.
% We validate our work with the more complex pixel-level classification task on a benchmark dataset VOC2012.
Experiments are conducted on four benchmark datasets, and our method acieves new state-of-the-art.

\end{abstract}
\vspace{-0.3cm}
\section{Introduction}
Deep neural networks have achieved great success in many tasks based on powerful computing devices and large-scale training data. However, deep learning suffers from catastrophic forgetting, i.e., when the network learns new tasks, its performance for the previous tasks dramatically degrades. To solve this problem, lifelong learning (also called continual learning or incremental learning) has been put forward, where the network can continually learn new tasks without forgetting the previous tasks.
Major approaches of lifelong learning can be categorized to regularization-based \cite{overcoming}, dynamic-architectures-based \cite{explicit} and replay-based \cite{icarl}.
Meanwhile, as artificial intelligence enters every aspect of people's lives, privacy protection and data leakage prevention receives more attention.
The new challenges include privacy-preserving localization \cite{localization}, learning from encrypted data \cite{learnfromdata}, and so on.
Lifelong learning cannot avoid this issue. Retaining the complete knowledge of all previous classes or tasks is a double-edged sword, which possibly leads to the risk of data leakage and invasion of privacy.

In order to solve the above mentioned problem, Learning with Selective Forgetting (LSF) is firstly proposed in \cite{lsf}, which aims to avoid catastrophic forgetting while selectively forget only specified sets of past classes. 
It designs a mnemonic code for each class to control which to forget and which to preserve, and only mnemonic codes belonging to preserved classes are used for the following training so as to enable the network to achieve selective forgetting.
Despite being a pioneer, this method leaves much to be desired.
Firstly, existing lifelong learning methods need to add an additional input (mnemonic codes) to achieve selective forgetting, but accurate generation of mnemonic codes is challenging for complex tasks like segmentation, which limits the usage.
% Although mnemonic codes have been proved harmless to the performance of image-level classification, it has great damage to other tasks like pixel-level classification.
Secondly, it indirectly affects the feature extraction through the loss of the classification head.
Intuitively, the direct operation of feature extraction is more effective and more universal (almost all deep learning has feature extraction part).
Thirdly, it just simply ignores the loss of deleted classes but does no specific forgetting operation on them, leading to forgetting very slowly and inefficiently, and affecting the accuracy of the preserved classes.
% leading to that the samples belonging to deleted classes can be classified to any class, which affects the accuracy of preserved classes.

Recently, contrastive learning shows significant advantages in lifelong learning \cite{sdr}, and based on this, we make an improvement to make the lifelong learning system easily obtain the ability of selective forgetting.
Specifically, for the preserved classes, we still follow the previous way of contrastive learning, i.e., making the aggregation of features within the class and the dispersion between the classes.
For the deleted classes, we make the features of different samples (features of pixels or images) within a same class dispersed and irregular so as to make the network react as if it is untrained to achieve selective forgetting.
Both forgetting and memory operate at the feature level (the embedding space), so they are very efficient and fast, and do not interfere with each other.

In conclusion, a more general strategy based on contrastive learning is proposed that perfectly incorporates lifelong learning and selective forgetting.
Moreover, the proposed method directly operates on the feature extraction part to make the forgetting fast and universal, and thus can fundamentally avoid the possibility of information leakage. 
Experiments on three classification and a segmentation benchmark datasets show the significant superiority of the proposed approach.
\vspace{-0.2cm}
\section{Related Work}
\subsection{Lifelong Learning}
Most of the current techniques of lifelong learning fall into three categories: regularization, dynamic- architectures, and replay-based.\\
\textbf{Regularization-based:} Regularization-based approaches are the most widely employed by far and mainly come in two flavours: penalty computing \cite{ewc} and knowledge distillation \cite{lwf,mib}.
Penalty computing approaches prevent forgetting via limiting the change of important weights inside the models.
Knowledge distillation relies on a teacher (old) model transferring knowledge to a student (new) model while the student model is still trained to learn new tasks.\\
\textbf{Dynamic-architectures-based:} Dynamic architectures grow new branches for new tasks, which can be explicit \cite{explicit}, if new network branches are grown; or implicit \cite{implicit}, if some network weights are available for certain tasks only.\\
\textbf{Replay-based:} Replay-based models exploit storing or generating examples during the learning process of new tasks.
Generative approaches \cite{generate} rely on generative models, which are later used to generate artificial samples to preserve previous knowledge.
Storage-based approaches \cite{icarl} save a set of raw samples of past tasks for the following training.
% during training for new tasks. 

Different from the pioneering work \cite{lsf} using both the penalty and distillation, ours is only categorized to the knowledge distillation-based approach, which contains simple structure and is more suitable for complex tasks such as segmentation \cite{mib,ssul}.

\subsection{Machine Unlearning}
Recent legislation such as the General Data Protection Regulation (GDPR) enacts the “right to be forgotten”, which allows individuals to request the deletion of their data by the model owner to preserve their privacy.
Moreover, any influence of the model about the target should also be removed. This process is referred to as machine unlearning (MU), which is first introduced by \cite{MU}.
General approaches are to train multiple small models on separated subsets of the training data to prevent retraining the whole knowledge \cite{Bourtoule} or to utilize vestiges of the learning process, i.e., the stored learned model parameters and the corresponding gradients \cite{Wu}.
Inspired by Differential Privacy (DP) \cite{Abadi}, Eternal Sunshine of the Spotless Net \cite{Golatkar} introduced a scrubbing procedure that removes knowledge from the already trained weights of deep neural networks using the Fisher information matrix.

To the best of our knows, we firstly employ latent contrastive learning for MU, and the operation is directly on the feature extraction part.

\subsection{Contrastive Learning}
Contrastive learning has recently become a dominant component in self-supervised learning methods \cite{contrastive_self_survey}. It aims at embedding augmented versions of the same sample close to each other while trying to push away embeddings from different samples.
In lifelong learning, SDR \cite{sdr} employs contrastive learning to cluster features according to their semantics while tearing apart those of different classes.
We follow the framework of SDR and make an effective improvement to endow it the ability of selective forgetting without affecting the lifelong learning.
\vspace{-0.2cm}
\section{Problem Definition}
% Learning with selective forgetting (LSF) is firstly proposed in \cite{lsf}, and we follow its setup.
$\left\{\mathcal{D}_{1}, \cdots \mathcal{D}_{t}, \cdots \mathcal{D}_{T}\right\}$ denotes a sequence of datasets, where $\mathcal{D}_{t}=\left\{\left(\mathbf{x}_{t}^{i},\mathbf{y}_{t}^{i}\right)_{i=1}^{n_{t}}\right\}$ is the dataset of the $t$-th task.
$\mathcal{S}_{t}=\mathcal{C}_{1} \cup, \dots, \mathcal{C}_{t}$, which denotes classes that have been studied.
We denote the input image space by $\mathcal{X} \in \mathbb{R}^{H \times W \times 3}$ with spatial dimensions $H$ and $W$, the set of classes (or categories) by $\mathcal{C}=\left\{c_{i}\right\}_{i=0}^{C}$, and the output space by $\mathcal{Y} \in \mathbb{R}^{H \times W}$ (i.e., the segmentation map), or $\mathcal{Y} \in \mathbb{R}^{1 \times 1}$ for classification tasks.
$\mathbf{x}_{t}^{i} \in \mathcal{X}$ is an input image, and $y_{t}^{i} \in \mathcal{Y}$ is the corresponding label.

In LSF problem, each of the learned classes is assigned to either preservation set or deletion set. 
\textbf{Preservation Set} $\mathcal{C}_{t}^{p}$ denotes a set of classes learned in the past and should be preserved at $t$-th task.
\textbf{Deletion Set} $\overline{\mathcal{C}_{t}^{p}}$ denotes a set of classes should be forgotten until $t$-th task (the complement of $\mathcal{C}_{t}^{p}$).
The objective of LSF is to learn a model $f_{\theta}: \mathcal{X} \rightarrow \mathcal{Y}$.
For the classes from current learning set or preservation set, the network is expected to predict correctly. Otherwise, for the deletion set, the network is expected predicts wrongly. 
Our work goes a step further and makes deleted classes to be classified as background classes (if there is) to eliminate the impact on the accuracy of preserved classes. \textbf{Note} that the network does not change architecture like deleting or merging the segmentation head.

% In \cite{lsf}, none of samples of the previously learned classes can be accessed when the new learning begins, which makes replay-based lifelong learning approaches impossible to achieve LSF.
% Although complete samples of the deleted classes cannot be acquired but the stored samples can still be used just like setting of the lifelong learning.
% In this way, the setting of lifelong learning and LSF is completely consistent and makes sense in more situations.
In lifelong learning, when new tasks start, samples of previously learned classes are not accessible \cite{lsf,mib}, and both \cite{lsf} and ours follow this rule.
However, the information of the old classes can be obtained indirectly, such as mnemonic code in the classification task, or the old class data obtained through pseudo-labels in the segmentation task, which can be used to achieve forgetting or memory.

\section{Proposed Method}
\subsection{Overview}
\begin{figure*}[htbp]
\centering
\includegraphics[scale=0.38]{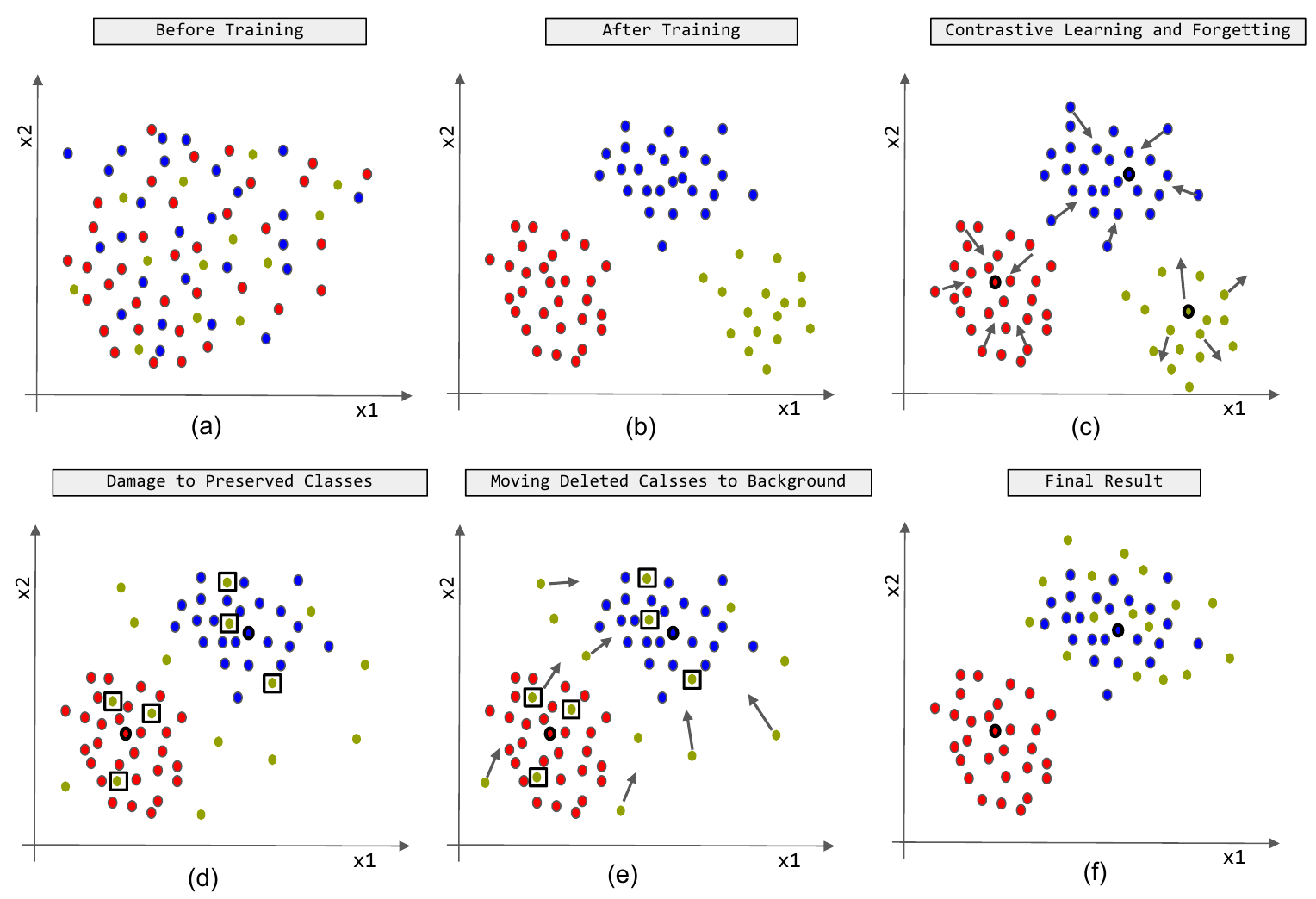}
\caption{Feature distribution and operations on features.
A color represents a class, and blue represents the background class.
A dot means the feature of a sample (an image or a pixel).
The features extracted from samples of the same class by untrained networks are scattered (shown in (a)), while the features from the trained networks are aggregated (shown in (b)).
Inspired by this phenomenon, we calculate the global prototypes for each class (the bold dots in (c)), and then diverge features belonging to the same deleted class (green dots in (c)), but compact features of the same preserved class (blue and red dots in (c)).
The above mentioned operations achieve selective forgetting but cause samples of deleted classes to be randomly categorized as any class, thus reducing the accuracy of preserved classes (the dots surrounded by black boxes in (d)).
We move features of deleted classes close to background class (shown in (e)), so the deleted classes are resembled to the background class (shown in (f)) to avoid affecting others.
}
\label{pic_intro}
\end{figure*}

\textbf{The Essence of Forgetting:}
If a network forgets a class, i.e., the network (including the feature extraction part and the classification part) does not contain any information about the class, thus no attack will lead to the leakage of information.
In conclusion, forgetting means that the features extracted by the network for different samples of a same class are scattered and irregular in all feature spaces.
Figure \ref{pic_intro} is vital to understand our idea, intuitively showing our motivation and the proposed operations.

Following SDR \cite{sdr}, we calculate one global (not limited to only one batch) prototype for each class, as shown in Section \ref{prototype}. 
Then, the prototypes are used to preserve and forget classes according to the mind of contrastive learning. Specifically, for the preserved classes, the network's response is expected to be consistent and compacted for samples of a same class;
for the deleted classes, our goal is to make the network respond randomly and irregularly as if the network has not been trained for this class at all. These procedures are introduced in Section \ref{contrastive_learn_forget}.
Meanwhile, for segmentation tasks, the features of deleted classes are moved close to background to avoid affecting the accuracy of other preserved classes, which is shown in Section \ref{background}.

\textbf{Symbol definition:}
We use step $t$ as the example, since the operations at different steps are consistent, we omit the flag $t$ for simplicity.
$\mathbf{y}^{\prime}_{n}= f_{\theta}\left(\mathbf{x}_{n}\right)$.
$f_{\theta}\left(\mathbf{x}_{n}\right)[h,w,c]$ is the probability for class $c$ in pixel $(h, w)$, where $h =1, \dots, H, w=1, \dots, W$.
And the output segmentation mask is computed as ${\mathbf{\tilde{y}}}_{n}=\arg \max _{c \in \mathcal{C}} f_{\theta}\left(\mathbf{x}_{n}\right)$.  
The background in ground truth ${\mathbf{y}_{n}}$ is replaced by the $\mathbf{\tilde{y}}_{n}$ of the previous step (step $t-$1), and the modified result is a comprehensive supervised information (pseudo label), called $\hat{\mathbf{y}}_{n}$.
The rationale behind this operation is that the prediction of old model to old classes is reliable, and the background class could incorporates statistics of previous classes.
Nowadays, for segmentation tasks, $f_{\theta}$ is typically an auto-encoder model made by an encoder $E$ and a decoder $D$ (i.e., $f_{\theta}=D \circ E$ ). We call $\mathbf{F}_{n}=E\left(\mathbf{x}_{n}\right)$ the feature map of $\mathbf{x}_{n}$, and
$\mathbf{y}_{n}^{*}$ is the downsampled comprehensive pseudo label (i.e., $\hat{\mathbf{y}}_{n}$) matching the spatial dimensions of $\mathbf{F}_{n}$.
For classification tasks, just change the pixel-level to image-level and pseudo label to mnemonic code to get the old class information, and the rest are the same.
% 对于分类任务，知识将pixel-level变为image-level，其余均一致。并且可以通过mnemonic code得到类似pseudo label一致的效果。

\subsection{Global Prototypes Calculation}
\label{prototype}

Prototypes (i.e., class-centroids) are vectors that are representative of each class, and we follow SDR to calculate the global prototype rather than prototype limited only to one batch.

For the current batch $\mathcal{B}$ with $B$ images, the current (or in-batch) prototypes $\hat{\mathbf{p}}_{c}[t]$ at $t$-th step are computed as the average of the features of class $c$, and the specific form is shown as,
\begin{equation}
\hat{\mathbf{p}}_{c}[t]=\frac{1}{B} \sum_{\mathbf{x}_{n} \in \mathcal{B}} 
\frac{\sum_{\mathbf{f}_{i} \in \mathbf{F}_{n}} \mathbf{f}_{i} \mathbbm{1}\left[y_{i}^{*}=c\right]}{\mid \mathbbm{1}\left[\mathbf{y}_{n}^{*}=c]|\right.},
\end{equation}
where $\mathbf{f}_{i} \in \mathbf{F}_{n}$ is a generic feature vector and $y_{i}^{*}$ is the corresponding pixel label in $\mathbf{y}_{n}^{*}$.
$\mathbbm{1}\left[\mathbf{y}_{n}^{*}=c\right]$ indicates the pixels in $\mathbf{y}_{n}^{*}$ associated to $c$, and when the label corresponding to the feature is class $c$, the value is 1; otherwise, the value is 0.
$|\cdot|$ denotes cardinality.

At training step $t$ with batch $\mathcal{B}$ of $B$ images, the global prototypes $\mathbf{p}_{c}[t]$ are updated for a generic class $c$ as:
% \begin{small}
\begin{equation}
\mathbf{p}_{c}[t]=\frac{1}{t}\left((t-1) \mathbf{p}_{c}[t-1]+ \hat{\mathbf{p}}_{c}[t] \right),
\end{equation}
% \end{small} 
which is initialized to $\mathbf{p}_{c}[0]=\mathbf{0}$ $\forall c$.
For classification tasks, an image has only one output, and the calculation method is the same.

\subsection{Lifelong Learning and Selective Forgetting}
\label{contrastive_learn_forget}
\textbf{Lifelong Learning:}
Contrastive learning is to structure the latent space to make features of the same class clustered near their prototype, which also helps in lifelong learning to mitigate forgetting and to facilitate the addition of novel classes. Formally, for the features within the same preserved classes, we define $\mathcal{L}_{in}^{p}$ to maintain the aggregation of in-class features, as follows:
\begin{equation}
\mathcal{L}_{in}^{p}=\frac{1}{N_c^p} \sum\limits_{
c_{j} \in (\mathcal{C}^{p}_{t} \cup \mathcal{C}_{t}) \atop c_{j} \in \mathbf{y}_{n}^{*} 
} \sum\limits_{\mathbf{f}_{i} \in \mathbf{F}_{n} } \| \left(\mathbf{f}_{i}-\mathbf{p}_{c_{j}}\right) \mathbbm{1}\left[y_{i}^{*}=c_{j}] \|_{F}\right.,
\label{L_cluster}
\end{equation}
where $N_c^{p}=\left|c_{j} \in \mathbf{y}_{n}^{*} \cap  c_{j} \in (\mathcal{C}^{p}_{t} \cup \mathcal{C}_{t}) \right|$, denoting the number of classes to be learned and preserved in $\mathbf{F}_{n}$.
We use the Frobenius norm $\|\cdot\|_{F}$ as metric distance.
$\mathcal{L}_{in}^{p}$ actually measures how close features are from their respective centroids, and its objective is to make feature vectors from the same preserved class are tightened around class feature centroids.
% $\mathbf{F}_{n}^{p}$ is the features of $\mathbf{F}_{n}$ and the corresponding pseudo label $\hat{\mathbf{y}}_{n}$ belongs to $\mathcal{C}^{p}_{t} \cup \mathcal{C}_{t}$.
The effect of $\mathcal{L}_{in}^{p}$ corresponds to the arrows surrounding red and blue dots in Figure \ref{pic_intro} (c).
For classification tasks, $\mathbf{F}_{n}$ is both the feature extracted from images (the feature of the current learning classes) and the feature extracted from mnemonic codes (the feature of the already learned classes).

\textbf{Selective Forgetting:}
On the contrary, if we want the network to forget a certain class, we can make the features within the class move away from the prototype at different distances and in all directions.
This requirement can be nicely satisfied, as shown below:

% \begin{footnotesize}
\begin{equation}
\mathcal{L}_{in, \mathbf{F}_{n}}^{d}=\frac{1}{N_c^{d}} 
\sum\limits_{c_{j} \in \overline{\mathcal{C}^{p}_{t}} \atop c_{j} \in \mathbf{y}_{n}^{*}} \sum\limits_{\mathbf{f}_{i} \in \mathbf{F}_{n}} 
\frac{1}{ \| \left(\mathbf{f}_{i}-\mathbf{p}_{c_{j}}\right) \mathbbm{1}\left[y_{i}^{*}=c_{j}\right] \|_{F}},
\end{equation}
% \end{footnotesize}
where $N_c^{d}=\left|c_{j} \in \mathbf{y}_{n}^{*} \cap  c_{j} \in \overline{\mathcal{C}^{p}_{t}} \right|$.
% $\mathbf{F}_{n}^{d}$ is the features of $\mathbf{F}_{n}$ but the pseudo label belongs to deleted classes $\overline{\mathcal{C}^{p}_{t}}$.
Furthermore, the distribution of same feature are different at diverse spaces. 
Based on this, we not only perform in-class dispersion in the last layer of feature extraction, but also add two $3 \times 3$ convolutions after the output to change its feature dimensions to the  $\frac{1}{2}$ and $\frac{1}{4}$ of original size (spatial size keeps same), and then carries out feature dispersion operation in these three feature spaces.
The two convolution can be thought of as a dimensional reduction operation, such that the feature is divergent from various embedding spaces. 
Therefore, $\mathcal{L}_{in}^{d}$ are computed as,
\begin{equation}
\mathcal{L}_{in}^{d}= \mathcal{L}_{in, \mathbf{F}_{n}}^{d}+\mathcal{L}_{in, \mathbf{F}_{n}^{*}}^{d}+\mathcal{L}_{in, \mathbf{F}_{n}^{**}}^{d},
\end{equation}
where $\mathbf{F}_{n}^{*}$ and $\mathbf{F}_{n}^{**}$ is the output of $\mathbf{F}_{n}$ after one and two $1 \times 1$ convolutions.
Note that the two additional features are only used to calculate $\mathcal{L}_{in}^{d}$ and do not enter the decoder part of the segmentation or the classifier of the classification networks.

\subsection{Making Deleted Classes Fall into Background}
\label{background}
Contrast learning can not only make the features within a same class close together, but also make different classes subject to a repulsive force, thus moving them apart.
We utilize $\mathcal{L}_{ex}^{p}$ to measure how far prototypes corresponding to different semantic classes are, i.e.,
% \begin{equation}
% \mathcal{L}_{ex}^{p}=\frac{1}{
% N^{p}_{c}
%  c_{j} \in (\mathcal{C}^{p}_{t} \cup \mathcal{C}_{t})
% \atop
% {c}_{j} \in \mathbf{y}_{n}^{*}}
% \sum_{c_{k} \in \mathbf{y}_{n}^{*} \atop c_{k} \neq c_{j}} \frac{1}{\|\hat{\mathbf{p}}_{c_{j}}-\hat{\mathbf{p}}_{c_{k}} \|_{F}}
% \end{equation}
\begin{equation}
\mathcal{L}_{ex}^{p} = \frac{1}{N_{c}^{p}} 
\sum_{
  \substack{
    c_{j} \in (\mathcal{C}_{t}^{p} \cup \mathcal{C}_{t}) \\
    c_{j} \in \mathbf{y}_{n}^{*}
  }
}
\sum_{
  \substack{
    c_{k} \in \mathbf{y}_{n}^{*} \\
    c_{k} \neq c_{j}
  }
}
\frac{1}{\|\hat{\mathbf{p}}_{c_{j}} - \hat{\mathbf{p}}_{c_{k}} \|_{F}}
\end{equation}

There are background in the segmentation task, i.e., others except the foreground classes are labeled as background.
A reasonable and feasible approach that prevent affecting the accuracy of the preserved classes is to move them close to the background to avoid false positives, as shown in Figure \ref{pic_intro} (e). The loss function to achieve this goal is shown as below,
\begin{equation}
\mathcal{L}_{ex}^{d}=\frac{1}{N_c^{d}} \sum\limits_{\mathbf{f}_{i} \in \mathbf{F}_{n}} 
\| \left(\mathbf{f}_{i}-\mathbf{p}_{b}\right) \mathbbm{1}[y_{i}^{*} \in \overline{\mathcal{C}_{t}^{p}}]\|_{F},
\end{equation}
where $\mathbf{p}_{b}$ is the prototype of background, and $N_c^{d}$ denotes the number of deleted classes in $\mathbf{F}_{n}$.

\subsection{Total Loss}
Consistent with SDR, to keep the effectiveness of global prototypes, we introduce $\mathcal{L}_{pc}$ to ensure the validity of the global prototype. 
More formally,
\begin{equation}
\mathcal{L}_{pc}=\frac{1}{\left|\mathcal{S}_{t-1}\right|}\left\|\mathbf{p}_{c}-\hat{\mathbf{p}}_{c}\right\|_{F} \quad c \in \mathcal{S}_{t-1}
\end{equation}

We modify the distillation loss from the whole previous learned classes $\mathcal{S}_{t}$ to only preserved classes $\mathcal{C}^{p}_{t}$, i.e.,
\begin{equation}
\label{ldis}
    \mathcal{L}_{dis}=-\frac{1}{W H} \sum_{w, h} \sum_{c \in \mathcal{C}^{p}_{t}} {y}^{\prime\prime}(w, h, c) \log {y}^{\prime}(w, h, c),
\end{equation}
where ${y}^{\prime\prime}$ is the output of the previous step (step $t-$1) model to keep the memory of already learned classes. 

Therefore, the training objective is summed as:
\begin{equation}
\mathcal{L}_{tot}=\mathcal{L}_{ce}+\mathcal{L}_{dis}+ \mathcal{L}_{pc}+\lambda_{p} \cdot (\mathcal{L}_{in}^{p}+\mathcal{L}_{ex}^{p}) +\lambda_{d} \cdot (\mathcal{L}_{in}^{d}+\mathcal{L}_{ex}^{d}),
\end{equation}
where the weights balance the multiple losses. 
The weights are the same as SDR, except $\lambda_{p}$ and $\lambda_{d}$. Unless specified, $\lambda_{p}$=$\lambda_{d}$=0.001.
And the results with different ratios of $\lambda_{p}$ and $\lambda_{d}$ have been shown in Section Sensitivity Analysis.
% The training objective is given by the combination of a modified cross-entropy loss $\left(\mathcal{L}_{ce}\right)$ with the above mentioned losses.
$\mathcal{L}_{ce}$ is the cross-entropy loss to learn new classes.
% 对于分类任务，其没有\mathcal{L}_{ex}^{d}损失。
For classification tasks, there is no $\mathcal{L}_{ex}^{d}$ loss.

% \begin{equation}
% \label{lpce}
%     \mathcal{L}_{ce}=-\frac{1}{W H} \sum_{w, h}^{W, H} \sum_{c \in \mathcal{C}^{p}_{t}} {y}(w, h, c) \log {y}^{\prime}(w, h, c)
% \end{equation}
\vspace{-0.2cm}
\section{Experiments}
\subsection{Settings}
\textbf{Dataset:}
For classification tasks, we use three widely used benchmark datasets for lifelong learning, i.e., CIFAR-100, CUB200-2011 \cite{cub}, and Stanford Cars \cite{stanford}. CUB200-2011 has 200 classes with 5,994 training images and 5,794 test images. CIFAR-100 contains 50,000 training images and 10,000 test images overall. Stanford Cars comprises 196 cars of 8,144 images for training and 8,041 for testing. 
Segmentation experiments are conducted on VOC (Pascal-VOC2012) \cite{pascal}.
The VOC contains 10,582 images in the training set and 1,449 in the validation set (that we use for testing, as done by all competing works because the test set not publicly available).
Each pixel of each image is assigned to one semantic label chosen among 21 different classes (20 plus the background). 

Cross-validation is used, i.e., 20\% of the training set is as the real validation set.
In line with \cite{lsf}, the first 30\% of classes for each task belongs to the deletion set, while the other classes belong to the preservation set.
If the number of the first 30\% classes is not an integer, we use method of rounding up.
We use the usually used overlapped setup for dataset division. In the first initial phase we select the subset of training images having only $\mathcal{C}_{0}$-labeled pixels. Then, the training set at each incremental step contains the images with labeled pixels from $\mathcal{C}_{t}$, i.e., $\mathcal{D}_{t} = \mathcal{X} \times \mathcal{Y}_{\mathcal{C}_{t} \cup\{b\}}$. Similar to the initial step, labels are limited to semantic classes in $\mathcal{C}_{t}$, while remaining pixels are assigned to $b$ (background).

\textbf{Implementation details:}
For the classification task, as in MC \cite{lsf}, we used ResNet-18 \cite{resnet} as the model. The final layer was changed to the multi-head architecture as in MC. The network is trained for 200 epochs for each task. Minibatch sizes are 128 in CIFAR-100, and 32 for CUB-200-2011 and Stanford Cars. The weight decay was $5.0 \times 10^{-4}$ and SGD is for optimization.
A standard data augmentation strategy is employed: random crop, horizontal flip, and rotation.
For segmentation task, we use the standard Deeplab-v3+ \cite{deeplabv3+} architecture with ResNet-101 \cite{resnet} as backbone with output stride of 16. 
SGD is used as optimiazation and with the same learning rate policy, momentum and weight decay as SDR. The first learning step involves an initial learning rate of $10^{-2}$, which is decreased to $10^{-3}$ for the following steps. The learning rate is decreased with a polynomial decay rule with power $0.9$.
In each learning step we train the models with a batch size of 20 for 30 epochs. The images are croped to $512 \times 512$ during both training and validation and the same data augmentation is applied, i.e., random scaling the input images using a factor from $0.5$ to $2.0$ and random left-right flipping.

Since the segmentation task can demonstrate the effectiveness of all our work, we conduct ablation experiments on the segmentation task.
The backbones are all been initialized using a pre-trained model on ImageNet \cite{imagenet}.
For the segmentation task, our project is based on \cite{sdr}, and for the classification task, the project is based on \cite{lsf}, i.e., we can indirectly obtain the information of the forgotten classes through mnemonic code or pseudo-labels in the background.
We use Pytorch to develop and train all the models on two NVIDIA 3090 GPUs. The new model is initialized with the parameters of the previous step model.

\textbf{Introduction of comparison methods:} We compare our proposed method with  the popular and state-of-the-art distillation-based lifelong learning methods. Further more, we also compare the modified version of them specifically for the LSF task.
To sum up, the specific methods compared are as follows:\\
% \begin{itemize}
- \textbf{FT}: Fine-tuning, trained using only the classification loss, which can be regarded as the upper bound of forgetting.\\
- \textbf{EWC} \cite{ewc}, \textbf{MAS} \cite{mas}, \textbf{LwF} \cite{lwf}, \textbf{MiB} \cite{mib}, \textbf{SSUL} \cite{ssul}, \textbf{SDR} \cite{sdr}, and \textbf{MC} \cite{lsf}.\\
- \textbf{LwF$^{*}$}, \textbf{EWC$^{*}$}, \textbf{MiB$^{*}$}, \textbf{SDR$^{*}$}, and \textbf{SSUL$^{*}$}: Modified version of LwF,EWC, MiB and SDR, i.e., the distillation losses or restricted parameters are changed from the calculation of all classes to only the preserved classes.\\
- \textbf{LE$^{*}$}: \text{LwF}$^{*}$+\text{EWC}$^{*}$.\\
- \textbf{LM$^{*}$}: \text{LwF}$^{*}$+MAS.\\
% \item[-] \textbf{MC}: Mnemonic Code (MC) \cite{lsf}, and the original method is used for classification task, we change it to pixel-wise classification. 
- \textbf{MC$_{E}$}: \text{MC}+\text{LwF}$^{*}$+\text{EWC}$^{*}$.\\
- \textbf{MC$_{M}$}: \text{MC}+\text{LwF}$^{*}$+\text{MAS}.
% And we employ the version which works best, i.e., the version using MAS \cite{mas} and LwF$^{*}$. We reproduce the code according to the original paper.
% \end{itemize}

\textbf{Evaluation metric:}
We follow the evaluation metric $S$ of \cite{lsf}, which called Learning with Selective Forgetting Measure (LSFM).
LSFM is calculated as the harmonic mean of the two standard evaluation measures for lifelong learning \cite{accuracy}: the average accuracy $A_{t}$ for the preserved classes and the forgetting measure $F_{t}$ for the deleted classes, i.e.,
\begin{equation}
S_{t}=\frac{2 \cdot A_{t} \cdot F_{t}}{A_{t}+F_{t}} .
\end{equation}
The average accuracy $A_{t}$ is evaluated after the model has been trained until the $t$-th task. The specific definition is given by $A_{t}=\frac{1}{t} \sum_{p=1}^{t} a_{t, p}$, where $a_{t, p}$ is the accuracy for the $p$-th task after the training for the $t$-th task is completed. $A_{t}$ is evaluated only for the preserved classes. Similarly, the forgetting measure $F_{t}$ is computed for the deleted classes after completing the $t$-th task. This measure is given by $F_{t}=\frac{1}{t} \sum_{p=1}^{t} f_{t}^{p}$, where $f_{t}^{p}=\max _{l \in 1 \cdots t} a_{l, p}-a_{t, p}$, which represents the largest gap (decrease) from the past to the current accuracy for the $p$-th task. This is evaluated only for the deleted classes. The ranges of $A_{t}$ and $F_{t}$ are both $[0,1]$.
We also report the averages of $S_{t}, A_{t}$ and $F_{t}$ after the last task has been completed, which are denoted by $S, A$, and $F$.\\

\vspace{-0.2cm}
\subsection{Results}
\begin{table}[H]
\centering
  \scalebox{0.75}{
   \begin{tabular}{c|cc|cc|cc}
   &\multicolumn{2}{c|}{CIFAR-100}&\multicolumn{2}{c|}{CUB-200-2011}&\multicolumn{2}{c}{Stanford Cars}\\
   &\multicolumn{2}{c|}{$\#$ Task:5, $\#$ Class:20}&\multicolumn{2}{c|}{$\#$ Task:5, $\#$ Class:40}&\multicolumn{2}{c}{$\#$ Task:4, $\#$ Class:49}\\
   &$S \uparrow$&($A \uparrow$,$F \uparrow$)&$S \uparrow$&($A \uparrow$,$F \uparrow$)&$S \uparrow$&($A \uparrow$,$F \uparrow$)\\
    \toprule
FT & 51.8 & (39.7, 74.6) & 42.2 & (31.8, \textbf{62.9}) & 48.1 & (41.1, 58.1)\\
\midrule
EWC& 48.6 & (36.5, 72.4) & 41.4 & (33.0, 55.5) & 48.7 & (47.7, 49.8) \\
$EWC^{*}$& 49.6 & (36.6, 77.1) & 42.1 & (33.4, 56.9) & 50.7 & (46.2, 56.2) \\
MAS& 47.5 & (34.9, 74.2) & 45.1 & (34.9, 63.7) & 48.8 & (44.7, 53.8) \\
\midrule
LwF& 17.2 & (79.1, 9.7) & 15.2 & (69.0, 8.5) & 9.9 & (88.1, 5.3) \\
$LwF^{*}$& 68.2 & (81.3, 58.8) & 44.5 & (68.3, 33.1) & 53.7 & (88.2, 38.6) \\
$LE^{*}$ & 67.6 & (81.2, 58.0) & 43.4 & (69.3, 31.6) & 52.8 & (88.7, 37.6) \\
\midrule
$LM^{*}$& 66.4 & (\textbf{81.8}, 55.8) & 47.5 & (\textbf{69.7}, 36.0) & 50.6 & (\textbf{89.0}, 35.3) \\
$MC_M$ & 73.2 &(72.6, 73.8) & 58.0&(63.1, 53.6) &72.2&{(84.6, 63.0)} \\
$MC_E$ & 79.6 & (75.3, 84.4) & 61.4 & (66.0, 57.4) & 73.7 & (86.0, 64.5)\\
\midrule
Ours&\textbf{81.7}&(78.2, \textbf{85.6})&\textbf{64.0}&(67.4, 61.0)&\textbf{75.9}&(87.2, \textbf{67.1})\\
  \bottomrule
\end{tabular}
}
\caption{The results (\%) on classification tasks.}
\label{tabel_result_cls}
\end{table}
\vspace{-0.2cm}
\textbf{Results on classification tasks:}
% 各种方法在分类任务上的结果如表1所示。
% 可以看出，如LwF对记忆类别非常好，但是对于遗忘类别表现较差，改进版本虽有改善但是依旧不理想。
% EWC和MAS，由于其本身的局限性，其对保留类别的记忆很差；相反地对删除类别的的性能有一定的优势，这不是因为其对unlearning具有优势，而是其遗忘了几乎所有的类别。
% LE和LM作为LwF与基于参数更新限制的结合，极大的弥补了LwF在遗忘能力上的缺陷，获得了非常显著的进步。
% MC作为专门为LSF任务设置的网络,相比于之前的方法，获得了巨大的进步。
% 我们的方法在特征空间上进行了组织，使得类别之间分散从而干扰减少。另外我们直接打散遗忘类别的特征分布，使得遗忘更完全。所以，我们在遗忘和记忆的综合性能上取得了最好的效果。
The results of various methods on classification tasks are shown in Table \ref{tabel_result_cls}. ‘Task:5, Class:20’ denotes that there are 5 tasks (steps) and 20 classes are learned per task.
It can be seen that LwF shows good performance on the memory for the preserved classes, but it performs poorly for the deleted classes. Although the modified version achieves some improvements, it is still not ideal.
EWC, MAS, and their modified versions, due to the own limitations, have a poor memory for preserved classes; on the contrary, they have certain advantages in the performance of the deleted classes.
This advantage is not of its advantage over unlearning, but because they forget almost all old classes.
LE$^{*}$ and LM$^{*}$, as the combination of LwF$^{*}$ and parameter update constraint-based methods, greatly make up for the shortcomings of LwF$^{*}$ in forgetting ability and achieve very significant progress.
As a network specially set up for LSF tasks, MC$_{E}$ and MC$_{D}$ make great progress compared to previous methods.
Our method organizes the feature (embedding) space so that the classes are dispersed, reducing the coupling between different classes.
In addition, we directly disperse the feature distribution of the deleted classes to make the forgetting more complete. Therefore, we have achieved the best results in the comprehensive performance of forgetting and memory.
\vspace{-0.2cm}
\begin{table}[H]
\centering
  \scalebox{0.78}{
   \begin{tabular}{c|cc|cc|cc}
   &\multicolumn{2}{c|}{19-1 (two tasks)}&\multicolumn{2}{c|}{15-5 (two tasks)}&\multicolumn{2}{c}{15-1 (six tasks)}\\
   &$S \uparrow$&($A \uparrow$,$F \uparrow$)&$S \uparrow$&($A \uparrow$,$F \uparrow$)&$S \uparrow$&($A \uparrow$,$F \uparrow$)\\
    \toprule
    FT&      27.2&(16.2, 85.6)&44.4&(29.6, \textbf{88.8})&16.4&(9.0, \textbf{89.9})\\
   
    MiB&     26.7&(82.8, 15.9)&8.7&(84.2, 4.6)&49.2&(40.3, 63.1)\\
    MiB$^{*}$& 26.3&(82.8, 15.6)&17.4&(83.7, 9.7)&52.5&(40.2, 75.7) \\
    SSUL &20.2&(\textbf{88.4}, 11.4)&19.0&(\textbf{86.2}, 10.7)&47.5&(\textbf{56.9}, 40.7)\\
    SSUL$^{*}$&21.7&(88.2, 12.4)&28.7&(81.6, 17.4)&48.1&(53.5, 43.7)\\ 
    SDR&     26.9&(79.6, 16.2)&22.5&(77.2, 13.2)&32.7&(20.0, 89.0)\\
    SDR$^{*}$& 34.8&(69.4, 23.2)&31.7&(75.4, 20.1)&31.9&(19.4, 89.2)\\
    MC$_{E}$&      51.0&(49.2, 53.1)&49.0&(40.1, 63.0)&24.7&(14.4, 86.9)\\
     LwF&     47.0&(61.0, 38.3)&35.2 &(66.6, 23.9)&24.2&(15.1, 61.0)\\
    LwF$^{*}$& 55.9&(58.9, \textbf{53.2})& 63.7&(66.7, 61.0)&26.6&(15.6, 89.4)\\
    \midrule
    Ours&\textbf{70.0}&(76.7, 50.6)&\textbf{72.9}&(67.1, 80.0)&\textbf{56.3}&(41.2, 88.7)\\
  \bottomrule
\end{tabular}
}
\caption{The results (\%) on segmentation tasks.}
\label{tabel_result_seg}
\end{table}
\vspace{-0.3cm}
\textbf{Results on segmentation tasks:}
Results on segmentation tasks \cite{shan2021class,shan2021decouple,shan2021densenet,shan2021uhrsnet,shan2022class,shan2022mbnet,shan2023boosting,shan2023data,shan2023incremental,wu2023continual,zhao2023explore,zhao2023flowtext,zhao2023generative,zhao2024controlcap} are shown in Table \ref{tabel_result_seg}. 
In line with incremental segmentation tasks, we set three learning rules, i.e., 19-1 (delete fist six classes), 15-5 (delete fist five classes), and 15-1 (delete fist five classes).
% If the number of the first 30\% classes is not an integer, we use method of rounding up.
15-1 is incremental learning one class in five steps, and the other two are similar.
% First of all, our proposed method is the best, and the superiority is significant.
FT achieves the best forgetting performance, but it forgets all the previous learned classes, resulting in low accuracy of preserved classes.
Compared with FT, LwF makes great progress in memory but loses much of forgetting capacity.
The forgetting performance of LwF$^{*}$ is significantly improved, and the memory capacity is not significantly decreased or even improved.
% SSUL
MiB and MiB$^{*}$ have strong memory capacity, but their forgetting capacities are poor in all three learning settings.
SDR, SDR$^{*}$, SSUL, and SSUL$^{*}$ are better than LwF in memory capacity, and better than MiB in forgetting capacity, so they can be regarded as a compromise version.
Although MC has a slight advantage in forgetting capacity, it has a great disadvantage in memory capacity, which because it adds mnemonic codes as extra information to the input during training, resulting in the inconsistent feature distribution between training and testing.
Our method is the best in terms of overall performance, and the superiority is significant, which demonstrates the advantages of contrastive learning for LSF tasks on segmentation tasks.

\vspace{-0.2cm}
\begin{table}[H]
\centering
   \scalebox{0.8}{
   \begin{tabular}{c|cc|cc|cc}
   &\multicolumn{2}{c|}{delete 5 classes}&\multicolumn{2}{c|}{delete 1 class}&\multicolumn{2}{c}{delete 10 classes}\\
   &$S \uparrow$&($A \uparrow$,$F \uparrow$)&$S \uparrow$&($A \uparrow$,$F \uparrow$)&$S \uparrow$&($A \uparrow$,$F \uparrow$)\\
    \toprule
    MiB$^{*}$&17.4&(83.7, 9.7)&40.0&(84.0, 23.7)&13.8&(85.3, 7.5)\\
    SSUL$^{*}$&28.7&(81.6, 17.4)&46.2&(82.4, 32.1)&27.2&(75.2, 16.6)\\
    SDR$^{*}$&31.7&(75.4, 20.1)&65.1&(71.6, 59.7)&31.1&(78.6, 19.4)\\
    LwF$^{*}$&65.5&(70.7, 61.0)&74.7&(67.5, 83.6)&58.5&(80.8, 45.9)\\
    \midrule
    Ours&    \textbf{72.9}&(67.1,80.0)&\textbf{80.8}&(74.2,88.7)&\textbf{66.4}&(81.6,56.0)\\
  \bottomrule
\end{tabular}
}
\caption{The results (\%) with varying deleted classes on setting of 15-5, and deleted classes are at front (sorted by class labels).}
 \label{table_dif_class}
\end{table}
% \begin{table}[H]
% \centering
%   \scalebox{0.8}{
%    \begin{tabular}{c|cc|cc|cc}
%    &\multicolumn{2}{c|}{10-10 (two tasks)}&\multicolumn{2}{c|}{10-5 (three tasks)}&\multicolumn{2}{c}{5-3 (six tasks)}\\
%    &$S \uparrow$&($A \uparrow$,$F \uparrow$)&$S \uparrow$&($A \uparrow$,$F \uparrow$)&$S \uparrow$&($A \uparrow$,$F \uparrow$)\\
%     \toprule
%     FT\\
% LwF$^{*}$&65.5&(70.7,61.0)&74.7&(67.5,83.6)&58.5&(80.8,45.9)\\
%     MiB$^{*}$&17.4&(83.7,9.7)&40.0&(84.0,23.7)&13.8&(85.3,7.5)\\
%     SDR$^{*}$&31.7&(75.4,20.1)&65.1&(71.6,59.7)&31.1&(78.6,19.4)\\
%     SSUL$^{*}$&31.7&(75.4,20.1)&65.1&(71.6,59.7)&31.1&(78.6,19.4)\\
%     \midrule
%     Ours&    \textbf{72.9}&(67.1,80.0)&\textbf{80.8}&(74.2,88.7)&\textbf{66.4}&(81.6,56.0)\\
%   \bottomrule
% \end{tabular}
% }
% \caption{The results (\%) on classification task.}
% \label{tabel_result}
% \end{table}
\textbf{Results with varying deleted classes:}
To give a more comprehensive picture to the performance of each method, we delete a different number of classes, and the results are shown in Table \ref{table_dif_class}.
Due to the inherent drawbacks of MC for segmentation, we do not compare it here.
When forgetting fewer classes like forgetting only one class, all methods perform well, indicating that forgetting a class is less difficult. When forgetting 10 classes, the forgetting performance of each method decreases significantly.
In all cases, our method shows the best performance.
 \vspace{-0.2cm}

\begin{figure}[H]
\centering
\includegraphics[scale=0.31]{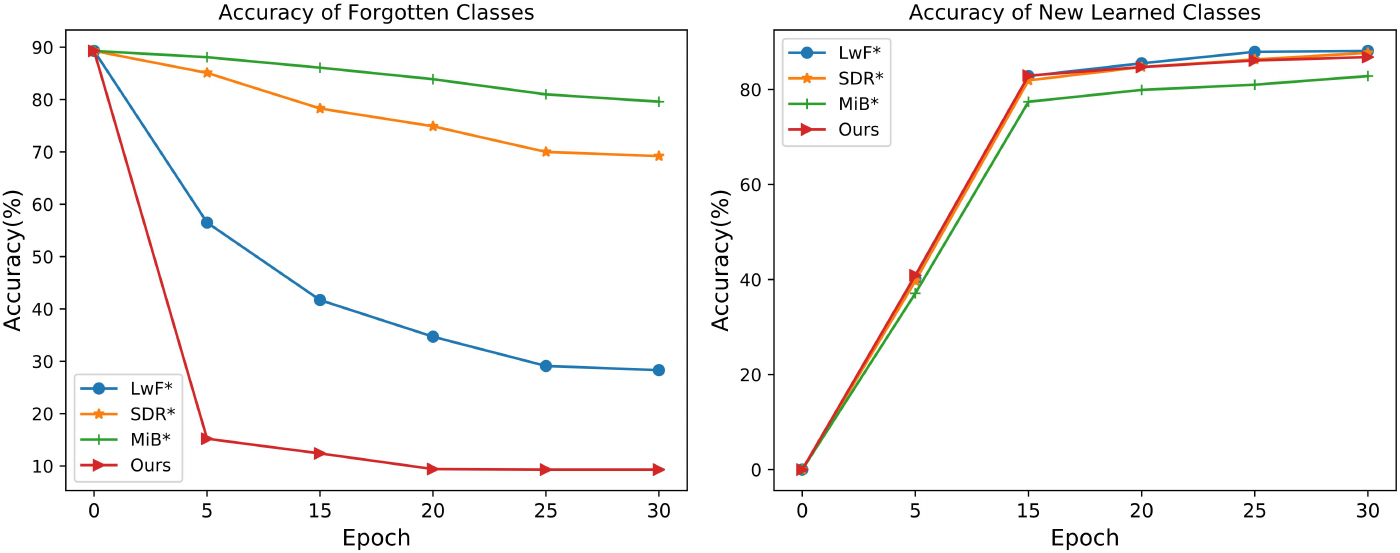}
\caption{The accuracy of deleted classes and new learned classes on different epochs with the setting of 15-5.}
\label{fig_dif_epoch}
\end{figure}
\vspace{-0.2cm}
\textbf{Per-epoch results:}
Figure \ref{fig_dif_epoch} shows the results during training.
As can be seen, all methods perform similarly to learn new classes (as shown on the right graph), but differ widely to forget classes.
Other methods (including LwF$^{*}$ MiB$^{*}$, and SDR$^{*}$) forget the deleted classes only through not calculating the loss of deleted classes, so the forgetting speed is very slow. These approaches are more accurately described as not learning rather than forgetting the deleted classes.
While our method makes targeted forgetting operation and directly act at the feature extraction part, so only 5 epochs are needed to reduce the accuracy of the deleted classes by 74.1 (89.3 to 15.2).
In conclusion, the way we propose to break up the features within the same deleted classes makes forgetting incredibly efficient and fast.

\begin{figure}[H]
\centering
\includegraphics[scale=0.31]{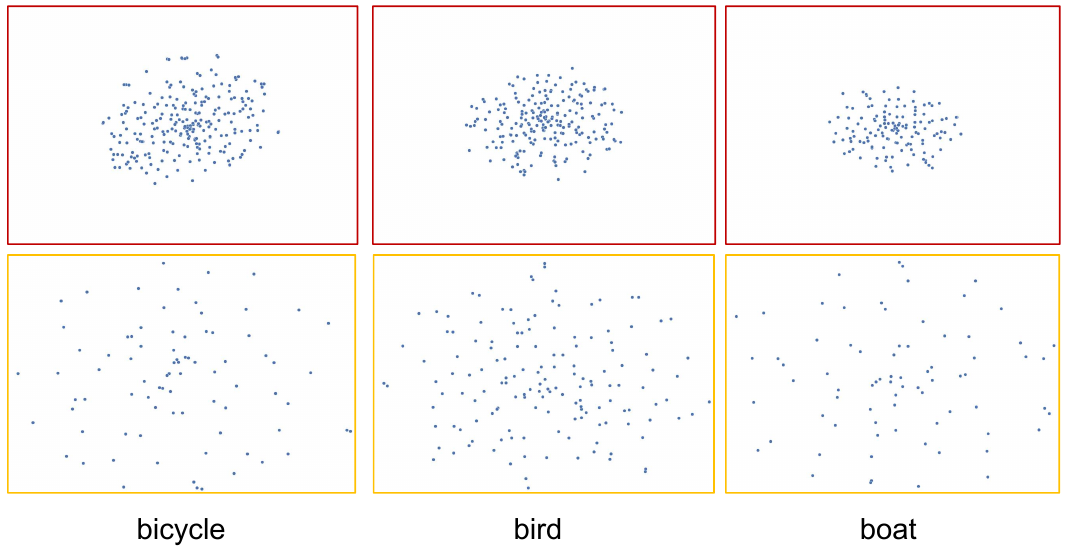}
\caption{
t-SNE plots of the features obtained from the last layer of the backbone before and after forgetting. 
Red boxes represents before forgetting and yellow is after, and the both are on the same scale.
Each point represents the feature of a pixel (after downsampling), and the whole graph is the visualization result of the features in a batch (batchsize=20). Bicycle, bird and boat are belonged to the deleted classes.
}
\label{fig_feature_show}
\end{figure}
\vspace{-0.2cm}

\textbf{Feature visualization:}
Figure \ref{fig_feature_show} shows the feature distribution before and after forgetting.
Since the classes are numerous and jumbled together, we only show the features of the deleted classes and show them separately to demonstrate the changes more clearly.
It can be seen that the features of the same class become divergent and irregular from the previous compacted situation after the forgetting operation.
% \newpage
\subsection{Ablation Study}
\vspace{-0.2cm}
\begin{table}[H]
\centering
  \scalebox{0.90}{
  \begin{tabular}{ccccc|c|c}
$\mathcal{L}_{pc}$&$\mathcal{L}_{in}^{p}$&$\mathcal{L}_{ex}^{p}$&$\mathcal{L}_{in}^{d}$&$\mathcal{L}_{ex}^{d}$&$S \uparrow$&($A \uparrow$, $F \uparrow$)\\
    \toprule
    % &&&&\\
    $\checkmark$&&&&&63.9&(67.1, 61.0)\\
    $\checkmark$&$\checkmark$&&&&60.4&(68.2, 54.2)\\
    $\checkmark$&$\checkmark$&$\checkmark$&&&61.0&(68.4, 55.1)\\
    \midrule
    $\checkmark$&&&$\checkmark$&&59.6&(47.4, 80.1)\\
     $\checkmark$&&&$\checkmark$&$\checkmark$&63.5&(52.2, 81.2)\\
     \midrule
    $\checkmark$&$\checkmark$&$\checkmark$&$\checkmark$&&71.0&(64.3, 79.4)\\  $\checkmark$&$\checkmark$&$\checkmark$&&$\checkmark$&61.2&(65.6, 57.4)\\
     \midrule
$\checkmark$&$\checkmark$&$\checkmark$&$\checkmark$&$\checkmark$&\textbf{72.9}&(67.1, 80.0)\\

    % $\checkmark$&$\checkmark$&&&\\
    \bottomrule
\end{tabular}
}
\caption{The results (\%) with varying deleted classes on 15-5.}
 \label{table_diff_losses}
\end{table}
\vspace{-0.2cm}

\textbf{The effectiveness of different losses:}
% 对不同损失的消融实验的结果如表1所示。
% 其中，Lpc是使得全局和局部的prototype保持一致，也是后面一系列操作的基础。
% lpin和lpex是用于保持对preserved classes的特征聚集来保持记忆的。可以看出，加上这两个损失之后，preserved classes的精确度有了部分提升，但是也导致对删除类别的遗忘不完全。
% 这可能是因为lpin和lpex增加了类内的聚合，而且又没有打散的操作导致网络无法遗忘旧类。
% ldin是保证遗忘的关键，其打散删除类别的特征分布从而达到遗忘的目的。可以看出，当加入ldin之后，遗忘性能大幅上升。但是，如果没有lpin和lpex，ldin也会导致preserved classes的性能下降。
% 而当lpin，lpex和ldin都存在时，保留类别的记忆和删除类别的遗忘均可以得到保证。因为这三者使得保留类别的特征分布保持了稳定，同时破坏了删除类别的特征分布，从而实现了LSF。
% 而当加上ldex之后，删除的类别不会影响保留的类别的性能，避免了False Positive，从而也提升了性能。
% 综上所述，所有的损失都发挥了作用，都是不可或缺的。
The results of the respective effectiveness of losses are shown in Table \ref{table_diff_losses}.
Among them, $\mathcal{L}_{pc}$ is to make the global and local prototypes consistent, and it is also the basis of the following series of operations.
$\mathcal{L}_{in}^{p}$ and $\mathcal{L}_{ex}^{p}$ are used to retain feature compaction of preserved classes to keep the memory. It can be seen that after adding these two losses, the accuracy of preserved classes has been partially improved, but it also leads to incomplete forgetting of deleted classes.
This phenomenon may be because $\mathcal{L}_{in}^{p}$ and $\mathcal{L}_{ex}^{p}$ increase the compaction within the class, and there is no disperse operation so that the network cannot forget the deleted classes.
$\mathcal{L}_{in}^{d}$ is the key to ensure forgetting, which breaks up the feature distribution of the deleted classes to achieve forgetting. It can be seen that when $\mathcal{L}_{in}^{d}$ is added, the forgetting performance increases significantly. Without $\mathcal{L}_{in}^{p}$ and $\mathcal{L}_{ex}^{p}$, $\mathcal{L}_{in}^{d}$ also causes performance degradation for preserved classes.
However, when $\mathcal{L}_{in}^{p}$, $\mathcal{L}_{ex}^{p}$ and $\mathcal{L}_{in}^{d}$ are all present, both the memory of preserved classes and the forgetting of deleted classes can be guaranteed.
LSF can be achieved due these three losses make the feature distribution of the preserved classes stable while destroying the feature distribution of the deleted classes.
After adding $\mathcal{L}_{ex}^{d}$, the results of deleted classes will not affect the performance of the preserved classes, avoiding false positives and thereby improving the performance.
In a nutshell, all losses play roles and are indispensable.

\begin{table}[H]
\centering
  \scalebox{0.86}{
   \begin{tabular}{ccc|c|c|c}
  $\mathbf{F}_n$&$\mathbf{F}_n^{*}$&$\mathbf{F}_n^{**}$&19-1&15-5&15-1\\
   &&&$S \uparrow$&$S \uparrow$&$S \uparrow$\\
    \toprule
      \checkmark&&&66.7&69.4&55.7\\
        \checkmark&\checkmark&&69.1&72.2&56.1\\
        \checkmark&\checkmark&\checkmark&70.0&72.9&56.3\\

    \bottomrule
\end{tabular}
}
\vspace{-0.2cm}
\caption{The results (\%) with divergence at different feature spaces.}
  \label{operation_space}
\end{table}
\vspace{-0.1cm}
\textbf{Effectiveness of divergence of different feature spaces:}
In order to completely break up the feature distribution of the deleted classes, in addition to performing contrastive forgetting on the features extracted and used for the final classification, we also add two convolutions to perform the dispersion in a higher-dimensional embedding space. The results of contrastive forgetting in different embedding spaces are shown in Table \ref{operation_space}. Adding $\mathbf{F}_n^{*}$ achieves an effect in all three different learning settings. However, after adding $\mathbf{F}_n^{**}$, the effect is not obvious and decreases in turn, which may be that the features of deleted classes have diverged enough in the embedding space to achieve forgetting. In a nutshell, the dispersion in the three embedding spaces all can bring improvements.

% \begin{table}[htbp]
% \centering
%   \label{table_1}
%   \scalebox{0.8}{
%   \begin{tabular}{ccc|c|c|c}
%   \multirow{2}{*}{$\mathbf{F}_{n}$}&\multirow{2}{*}{$\mathbf{F}_{n}^{*}$}&\multirow{2}{*}{$\mathbf{F}_{n}^{**}$}&19-1&15-5&15-1\\
%   &&&$S \uparrow$&$S \uparrow$&$S \uparrow$\\
%     \toprule
%     &&&&\\
%     &$\checkmark$&&&&\\
%     &&$\checkmark$&&&\\
%     &$\checkmark$&$\checkmark$&&&\\
%     \bottomrule
% \end{tabular}
% }
% \caption{The results (\%) of VOC2012 with varying deleted classes}
% \end{table}
% \textbf{Different Level Features:}
% Overall Results: Table 1 shows the comparative results of all the methods. We can clearly see that Ours $_{E}$ is the best and Ours $_{M}$ is the second best among all the methods in $S_{F}$. No other method that is better in terms of both $A$ and $F$. This is mainly due to the advantage of our mnemonic codes; as we verified in our preliminary analysis, the codes enable accurate control over whether each class should be retained or forgotten on a class-by-class basis.\\

% \vspace{-0.2cm}
\subsection{Sensitivity Analysis}
% \vspace{-0.2cm}
\begin{figure}[H]
\centering
\includegraphics[scale=0.34]{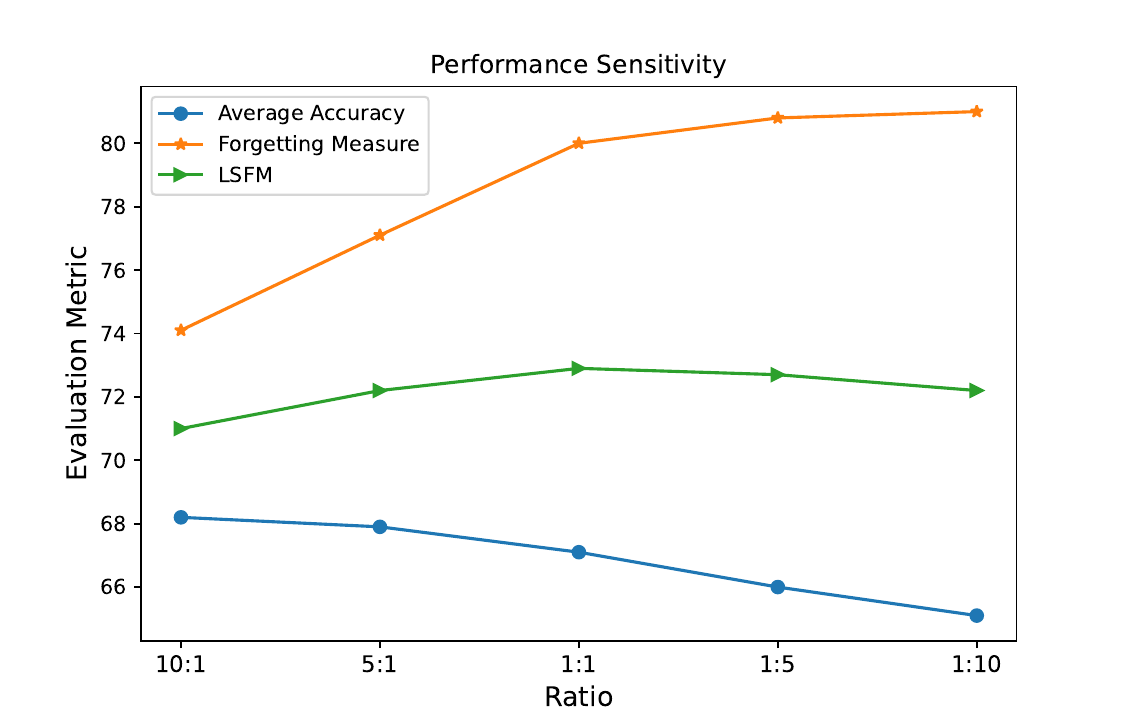}
\caption{Sensitivity analysis. The abscissa represents the ratio of $\lambda_{p}$ to $\lambda_{d}$.}
\label{fig_sensitivity}
\end{figure}
\vspace{-0.2cm}
Memory of preserved classes and forgetting of deleted classes are two different goals, so it is of great significance to explore the weight of  memory loss ($\mathcal{L}_{in}^{p}$ and $\mathcal{L}_{ex}^{p}$) and forgetting loss ($\mathcal{L}_{in}^{d}$ and $\mathcal{L}_{ex}^{d}$), i.e., $\lambda_{p}$ and $\lambda_{d}$.
Consistent with SDR, the losses calculated from features are multiplied by 0.001, i.e., 1:1 is actually 0.001:0.001.
The experimental results are shown in Figure \ref{fig_sensitivity}.
$\lambda_{p}$ corresponds to memory and $\lambda_{d}$ to forgetting. When $\lambda_{p}$ is large, the average accuracy of preserved classes is high, but the forgetting measure is relatively poor. When $\lambda_{p}$ gradually becomes smaller in the ratio ($\lambda_{d}$ becomes larger and gradually dominates), the forgetting measure becomes better, but the accuracy decreases.
This phenomenon is consistent with the previous analysis. When $\lambda_{d}$ is small, the distribution of the deleted classes cannot be broken up, so complete forgetting cannot be achieved.
When $\lambda_{p}$ is small, breaking up the forgotten category will affect the accuracy of the preserved classes and thus cause errors.
Therefore, the best effect is obtained when the two are 1:1.

% \vspace{-0.2cm}
\section{Conclusion}
In this paper, an LSF method based on a contrastive strategy is proposed to perform direct forgetting by computing the dispersion loss for the deleted classes, which destroys the distribution of the deleted classes to quickly achieve selective forgetting without affecting the feature distribution of the preserved classes. The effectiveness of the proposed method is demonstrated in classification and segmentation tasks. Besides, the forgetting operation is performed in the feature extraction part, so it can be easily extended to other tasks (almost all deep learning-based models need to extract features).
Our approach has implications for the continual and safe use of deep learning-based software in practical applications.
In future works, how to better deal with the features of the deleted classes that have been dispersed to prevent false positives is a problem that needs to be solved.
Similarly, exploring the learning ability and interpretability of neural networks from the aspect of feature distribution rather than the network structure is also a direction worth exploring.

% \newpage

\bibliographystyle{named}
\bibliography{ijcai22}

\end{document}